\pdfoutput=1
\documentclass[11pt]{article}

\usepackage[preprint]{latex/acl}

\usepackage{times}
\usepackage{latexsym}

\usepackage[T1]{fontenc}
\usepackage[utf8]{inputenc}
\usepackage{amsmath}
\usepackage{microtype}

\usepackage{inconsolata}

\usepackage{graphicx}
\usepackage{pifont} % For checkmark and X symbols
\usepackage{xcolor}
\definecolor{darkgreen}{rgb}{0,0.5,0}

\usepackage{bbm}
\title{Multi-LLM QA with Embodied Exploration}
\author{Bhrij Patel \\ University of Maryland, \\ College Park \And Vishnu Sashank Dorbala \\ University of Maryland, \\ College Park \\ \\ \textbf{Dinesh Manocha} \\ University of Maryland, \\ College Park\\ \And  \textbf{Amrit Singh Bedi} \\ University of \\ Central Florida  }
\begin{document}

\maketitle

\begin{abstract}
% Large Language Models (LLM) with their pretrained knowledge have seen rapidly increasingly success in question-answering (QA) tasks. More recently, a mixture of experts (MoE), is growing paradigm where multiple 
Large language models (LLMs) have grown in popularity due to their natural language interface and pre trained knowledge, leading to rapidly increasing success in question-answering (QA) tasks. More recently, multi-agent systems with LLM-based agents (Multi-LLM) have been utilized increasingly more for QA. In these scenarios, the models may each answer the question and reach a consensus or each model is specialized to answer different domain questions. However, most prior work dealing with Multi-LLM QA has focused on scenarios where the models are asked in a zero-shot manner or are given information sources to extract the answer. For question answering of an unknown environment, embodied exploration of the environment is first needed to answer the question. This skill is necessary for personalizing embodied AI to environments such as households. There is a lack of insight into whether a Multi-LLM system can handle question-answering based on observations from embodied exploration. In this work, we address this gap by investigating the use of Multi-Embodied LLM Explorers (MELE) for QA in an unknown environment. Multiple LLM-based agents independently explore and then answer queries about a household environment. We analyze different aggregation methods to generate a single, final answer for each query: debating, majority voting, and training a central answer module (CAM). Using CAM, we observe a $46\%$ higher accuracy compared against the other non-learning-based aggregation methods. We provide code and the query dataset for further research. 
\end{abstract}


\section{Introduction}

Large Language Models (LLMs) \citep{achiam2023gpt} become more embedded into society, with one of the most popular tasks being question-answering (QA) \citep{rawal2024cinepile, kamalloo-etal-2023-evaluating}. Further research has investigated utilizing an ensemble of LLM models (Multi-LLM) for the ability to discuss \citep{li2023prd, chen2023reconcile}, answer questions from large data sources and route questions to models with specialized knowledge (Mixture of Experts) \citep{li2024more}. 

However, in these prior works on ensemble LLMs for QA, the questions are asked in a zero-shot manner or information is \textit{provided} to the models. For QA of a physical environment, an embodied agent must first \textit{explore} an environment, such as a household, to answer questions based on its observations. It has gained considerable interest in the past few years due to its applications for in-home robots and personalized assistants \cite{das2018embodied, gordon2018iqa, duan2022survey, sima2023embodied, luo2023robust, dorbala2024s, zhu2023excalibur}. Possessing zero-shot common-sense reasoning of relational understandings of household items is crucial for LLMs to perform embodied exploration in unknown environments \cite{dorbala2023can, dorbala2024right, OpenEQA2023}. 

% For a seamless interface with humans, robotic agents must be equipped with natural language understanding \cite{xu2021grounding, sun2024beyond} and possess common-sense reasoning of the world. One form of common sense is having a relational understanding of household items and their usual place (i.e., a mug is highly likely to be in the kitchen or living room). 

% Foundation models, specifically Large Language Models (LLMs) such as GPT \cite{achiam2023gpt} and LLama \cite{touvron2023llama}, have shown remarkable prowess in both natural language understanding and common sense reasoning. 

%For an LLM-based embodied agent to succeed in EQA, efficient exploration of the environment is needed as the questions will center around the state of the environment \cite{das2018embodied, yu2019multi, zhu2023excalibur, OpenEQA2023}. Prior work has studied the zero-shot capabilities of LLM-based agents to explore an unknown household environment \cite{dorbala2023can, dorbala2024right, OpenEQA2023}. 
\begin{figure*}[ht]
    \centering
    \includegraphics[width=\textwidth]{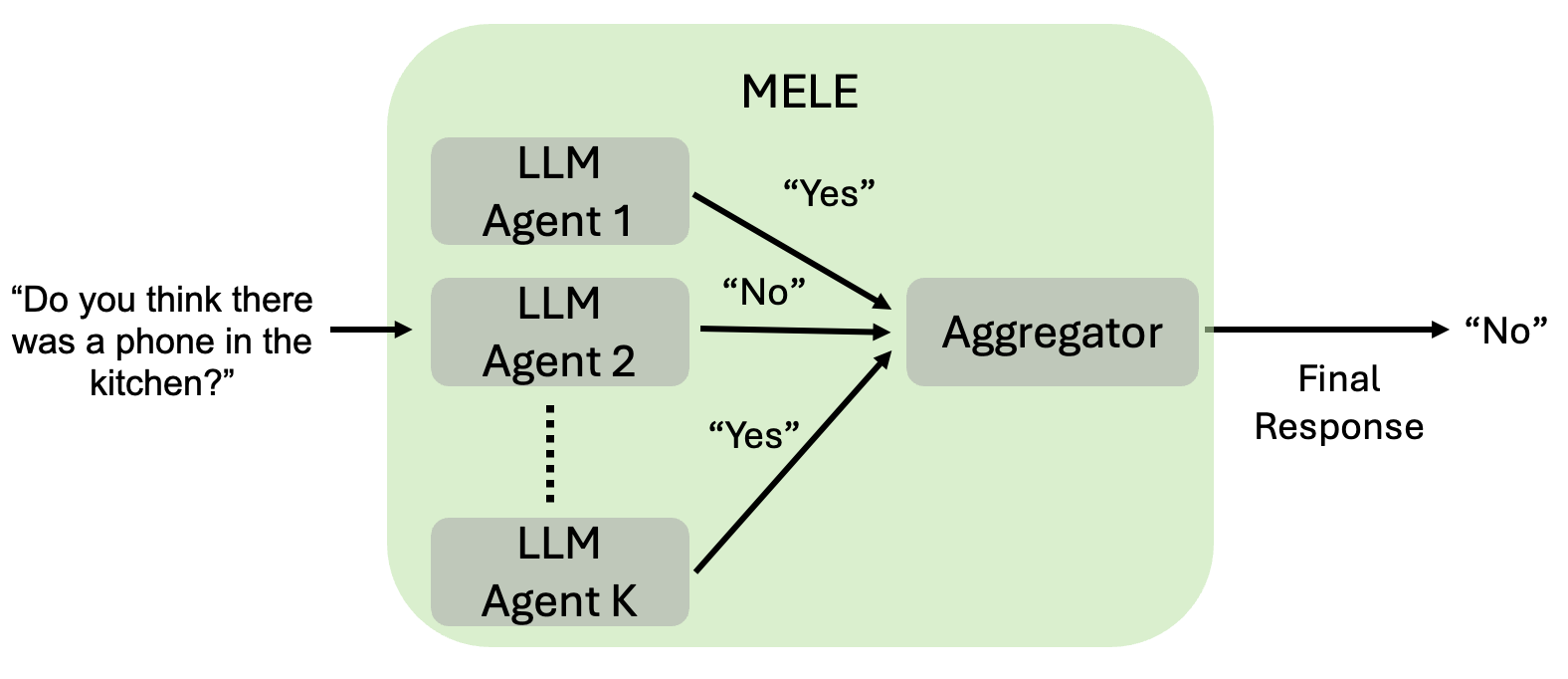}
    \caption{Overview of our MELE framework. LLM-based agents that have explored the dynamic household are independently asked a binary ``Yes/No'' question and must answer based on their observations. Only one of the agents has the correct answer (``No'') while the other two agents are wrong. The aggregator function then takes in these responses to decide a final response.}
    \label{fig:central_network}
\end{figure*}
In existing works, only a single agent explores the environment with a limited field of view, inducing low coverage and high exploration costs. In contrast, for an LLM-based embodied agent to succeed in EQA, efficient exploration of the environment is needed as the questions will center around the state of the environment \cite{das2018embodied, yu2019multi, zhu2023excalibur, OpenEQA2023}. The rise of Embodied Artificial Intelligence (EAI) \cite{eaisurvey} brings up the possibility of co-habited spaces with multiple agents and humans working in conjunction \cite{habitat30}. 

With this trend, we believe investigating the abilities of multiple embodied LLM agents to answer questions on their environment becomes increasingly crucial for the personalization of AI. We consider a similar co-inhabited setup and propose using a team comprising multiple embodied LLM-based explorers. An ensemble system of explorers allows for simultaneous exploration of the environment in the same amount of time.

However, there are also challenges when introducing a multi-agent setup. A key challenge in a multi-agent setup is with conflicting responses when agents are asked the same query. Since LLMs are blackboxes capable of generating free-form natural language, their responses to the same query may significantly vary \cite{multisur1, multillm1}. In an embodied setting, such responses are further skewed by partial observations due to limited exploration. Prior work in ensemble LLM methods tackle this issue using weighted sum or voting schemes \cite{li2024more, wei2022chain, wang2022self, yang2024llm} and consensus-reaching debates between the LLM agents \cite{chen2023multi, du2023improving, xiong2023examining, guo2024large, chan2023chateval, li2023prd}. However, such methods can be vulnerable to scenarios where incorrect answers from malicious or poor-performing agents can influence the overall decision \cite{amayuelas2024multiagent}. 

% Incorrect answers can stem from agents that have not explored relevant areas of the house or are malicious.
% An ideal EQA model would learn to identify incorrect agents and avoid being influenced by their responses. Furthermore, we would like to ideally avoid the communication costs incurred by debate-based approaches.

% using a consensus-reaching debate approach for each question during inference is inefficient due to the communication costs.

% Unreliability (HOW DO YOU DEFINE UNRELIABLITY HERE) can stem from agents that have not explored relevant areas of the house for the specific question or are malicious. 

% However, such methods can be vulnerable to scenarios where multiple unreliable (WHY ARE THEY UNRELIABLE) agents can influence the overall decision. Unreliability (HOW DO YOU DEFINE UNRELIABLITY HERE) can stem from agents that have not explored relevant areas of the house for the specific question or are malicious. Furthermore, having the LLM agents debate WHY DOES THE ISSUE OF DEBATE ARISES with one another for each question during inference is inefficient.

% \am{once you have mentioned the challenges, say C1, C2, and C3, then in this paragraph you describe your approach to address them on high level and then summarize the contribution.} 
\textbf{Main Results:} To investigate the use of ensemble LLM for QA with embodied exploration, we apply a system of Multi-Embodied LLM Explorers (MELE) to solve this embodied task. We instruct the agents to answer based on their observations a set of binary embodied questions about the household. We then use various aggregation methods to output a final answer based on the independent responses from the agents: majority voting, debating, and training a supervised central answer model (CAM). Majority voting and debating are common aggregation methods in existing literature and we propose investigating training CAM for more robustness to adversarial or incorrect agents.
 % The network directly outputs the final answer to a given question and does not require any communication between agents. Furthermore, our model learns which agents to rely on while producing its final answer, which reduces the end user's effort in having to individually figure out the best agent response. 
% \textbf{Main Results:} To address the issues presented above, we solve EQA using a novel Multi-LLM agent approach. We instruct the agents to answer based on their observations a set of binary embodied questions about the household. We then train a Central Answer Model (CAM) on the answers of independent agents with supervised learning. The network directly outputs the final answer to a given question and does not require any communication between agents. Furthermore, our model learns which agents to rely on while producing its final answer, which reduces the end user's effort in having to individually figure out the best agent response. 
% \am{contributions are not crisp, can you do something like short boldface summary in the starting for each point  as well ?}
Our novel contributions include:

\begin{itemize}

    \item \textbf{Multi-LLM QA Framework with Embodied Navigation:} We use multiple independent embodied LLM agents that explore the environment individually generate a response and use various aggregator functions to predict a final response.  \footnote{Link to anonymous repository: \url{https://anonymous.4open.science/r/CAM-embodied-llm-54CE/}} We depart from prior Embodied Question-Answering (EQA) \cite{das2018embodied} setups where the agents explore after being asked a question. Rather we have the agents explore before being asked a set of questions on their environment. We generate queries for various Matterport 3D \cite{matterport3d} environments. We train CAM on the labeled query dataset with multiple machine learning methods and show that the best CAM-based method can achieve up to $\textbf{46\%}$ higher accuracy than a majority vote and debate aggregation methods. While prior work has used a central answer module to output a final answer from a mixture of answers from multiple LLM experts \cite{puerto-etal-2023-metaqa, si-etal-2023-getting, jiang-etal-2023-llm}, we are the first to apply the framework to embodied agents that must first explore an environment. 

    \item \textbf{Integration with Exploration Systems:} We evaluate our multi-agent framework on data gathered using an LLM-based exploration method \cite{dorbala2023can} on multiple agents in Matterport3D, demonstrating our system can work in conjunction with SOTA exploration methods in unknown environments. 
    
    \item \textbf{Examining Agent and Feature Reliance:} We present a feature importance analysis of CAM reliance on each independent agent using the metric, permutation feature importance (PFI) \cite{breiman2001random, Fisher2018AllMA}. We also look at the agreement between the individual agent and the final answer of the central answer model and the baselines. Similarly to \citet{si-etal-2023-getting}, some of our methods for the central answer model are interpretable and provide a Decision Tree visualization of one of our models.

    % \item We are the first to tackle EQA in a dynamic environment with non-stationary items DO YOU MEAN OBSTACLES.
    % \item We show that a Multi-LLM system can be better than a single LLM agent in terms of accuracy because more observations can be observed in the same amount of time. IS THIS EMPIRICAL OBSERVATION OR YOU PROVE IT?
    % \item Training a central classifier is as least as accurate than having the multiple LLM agents communicate with each other during inferencing. Having the LLM agent needing to reach a consensus for each question is inefficient. Our approach does not require communication.
    % \item We show the robustness of this Multi-LLM framework against unreliable agents (ones that have not explored enough, hallucinate, or are intentionally wrong) as the central classifier learns to weigh the answers of the different agents. We also show that the reliability of an agent depends on the question and is connected to how much of the house it explores and for how long. For example, if one agent has mainly stayed in the kitchen, the central network will learn to weigh that agent’s answers more when faced with a kitchen question.
 
    % Training a central network to give a final answer helps with robustness against unreliable agents (ones that haven’t explored enough, hallucinate, or are intentionally wrong). Or if one agent has mainly just stayed in the kitchen, the central network will learn to weigh that agent’s answers more when faced with a kitchen question. We can compare against doing a majority vote of the independent answer.
\end{itemize}

\section{Related Works}\label{sec:related_works}

\subsection{Embodied Question Answering and Exploration}

Embodied Question Answering (EQA) was first introduced by \citet{das2018embodied} and has become a popular task for robotics \citep{gordon2018iqa, gordon2019should, duan2022survey, sima2023embodied, wijmans2019embodied, chen2019audio, luo2019segeqa}, where traditionally an agent is given a query such as ``was there a plant in the living room?' and explores its environment, in this case a house, to answer the question. \citet{zhu2023excalibur} had alternated between an exploration phase and then a question phase and repeated the process until the query was answered. Further work has extended EQA to scenarios where the final answer revolves around multiple objects \cite{yu2019multi}, and where a target object is not explicitly stated in the question \cite{tan2023knowledge}. \citet{tan2020multi} introduced multi-agent exploration and cooperation for question-answering in interactive environments. \citet{dorbala2024s} provided a dataset of subjective or situational queries and was the first to use a generative approach for query creation, and \citet{OpenEQA2023} designed the first open-vocabulary EQA dataset to test and benchmark foundation models on this task. Recently, \citet{singh2024evaluating} studied the zero-shot LLM performance on visual QA with indoor house scenes. However, the LLM is given the scenes without any exploration of the environment. We evaluate our Multi-LLM approach with a SOTA LLM-based exploration method for observation gathering introduced by \citet{dorbala2023can}.

\textbf{Note:} In our work, we deviate from the traditional EQA approach. We first have an exploration gathering phase, then answer a set of questions in the same environment. We do so for two reasons: 1) to provide a setup that better resembles prior work in LLM-based MoE QA \citep{li2024more} 2) we believe this setup is better suited for the scenario where the user asks a question about the items in their homes and quickly needs an answer.
% To the best of our knowledge, we are the first work to analyze the effect of the exploration behavior of an LLM on EQA task performance. Furthermore, the prior art assumes access to a single embodied LLM agent. In this work, we develop the first multi-LLM framework for EQA using a dynamic environment with non-stationary items developed by \citet{dorbala2024right}. 

\subsection{Systems of Multiple LLM-based Agents}

Research in LLM-based systems have grown in popularity with the rise of foundation models for areas such as task planning \cite{chen2023scalable}, application development \cite{wu2023autogen}, and chatbots \cite{duan2023botchat}. Extensive studies have focused on their abilities to cooperate. Prior work has used debate between LLM models to reach a consensus \citep{chan2023chateval, du2023improving, guo2024large, chen2023reconcile}. \citet{chen2023multi} analyzes the consensus-seeking debate between multiple LLM agents with different personalities, and \citet{amayuelas2024multiagent} investigates the susceptibility of a LLM-based discussion to an adversarial agent that purposefully tries to lead an incorrect answer choice. For embodied AI, \citet{guo2024embodied} have shown that multiple LLM-based agents can cooperate in teams by designating a leader for increased task efficiency. Other work have looked into using voting metrics of independent LLM-agents to arrive at a final answer \citep{li2024more, wei2022chain, wang2022self}. Our work aims to remove the need for communication by training a central classifier on independent answers from multiple agents. Training a classifier in a supervised manner helps our approach become robust to unreliable agents that may influence overall answers in voting schemes.

A central architecture to output a final answer from multiple answers from different LLM experts has been explored previously \cite{puerto-etal-2023-metaqa, si-etal-2023-getting, jiang-etal-2023-llm}. Our approach differs from these prior works as the various LLM agents we utilize become experts in the environments from an embodied exploration phase and we show it can be used with off-the-shelf exploration methods. Furthermore, \citet{wang2024mixture} introduced the idea of a Mixture of Agents to enhance task performance of LLMs. We specifically look at a mixture of embodied agents that explore a unknown environment. 
\section{Multi-Embodied LLM Explorers}\label{sec:pipeline}

\subsection{Problem Formulation} \label{sec:problem_formulation}
In this section, we provide a mathematical formulation of our specific EQA setting to characterize the EQA problem we are tackling in this work. We consider a tuple of $(Q, S^K)$, where $Q$ represents the set of queries, where each sample is a vector $\left[o, r, y\right]$ with object $o$ and room $r$ representing the question ``Is there an $o$ in the $r$?'', and $y \in \{ 0,1 \}$ is the binary ground-truth answer to the question. $S$ is the set of $K$ embodied agents that independently answer the questions. Let $s_k^{(o,r)} \in \{ 0,1 \}$ be the independent answer to query $(o,r)$ of the $k$-th agent. Let $f$ be a aggregation function such that $f\left(o, r, \cup_{k=1}^K s_k^{(o,r)}\right) = \hat{y} \in \{ 0,1 \}$. Given the query context $(o,r)$ and the independent agent answers to the context, $f$ should produce a final binary prediction of whether object $o$ is in room $r$. 

\subsection{Our Approach}
We implement MELE as a two-step procedure that comprises a \textit{question-answering phase}, and a \textit{aggregation phase}. The question-answering phase is an agent-specific step while the classification-training step relies on combining the intermediate results of the question-answering phases from various agents. Before starting the question-answering phase, the agents explored the environment to gather observations. Later, we examine the performance of MELE when using a LLM-based SOTA method for exploration. 

\subsection{Question-Answering Phase}\label{sec:qa}

In this phase, we generate a dataset of binary embodied questions specific to a Matterport 3D environment. Matterport3D is a large-scale dataset popular for scene-understanding tasks and embodied AI research, containing RGB-D panoramic views of various building environments like homes. The data is stored as a graph for each environment where the nodes are different physical locations within the home environment that an embodied agent can occupy. We use a topological graph environment because we are only concerned with high-level exploration policy in the house environment. Our approach can be used with other similar environments such as HM3D \cite{mp3dhabitat}.

Each LLM-based agent's answers are based on independent observations from its exploration of the environment. To emphasize, \textbf{the agents differ in their collected observations, leading to differing answers of the same query.} To generate these questions we follow this template:
 
% \textit{Was there a [OBJECT] in the [ROOM] at [TIMESTEP]?}
\textit{Do you think there is a [ITEM] in the [ROOM]? Use both common-sense reasoning about the object and room and the observation list given. Even if the relevant information is not in your observations, the answer can still be YES. Respond with YES or NO.}
 
We then prompt the LLM to answer each question based on the stored observations of the agent. We pass the observations to the system prompt formatted as a dictionary, \{\textit{ROOM NAME}: [\textit{LIST OF ITEMS IN THE ROOM}]\}. The user prompt is a question following the template above and the system prompt is as follows: \textit{You are an embodied agent that has explored a house. I prefer definite answers to questions. The observations are: [OBSERVATION DICTIONARY]}

% We also specify sometimes for the LLM to purposefully give wrong answers to simulate malicious agents. We then store the answers offline.

\subsection{Aggregation Phase}
We now describe the different methods used for aggregation function $f$. 

\subsubsection{Non-learning-based Aggregation}
We first describe the non-learning based methods used commonly in ensemble LLM literature. 
\begin{itemize}
    \item \textbf{Majority Vote (MV):} For each test question, we take the majority vote of the independent answers from the agents of the Multi-LLM system. We do not directly take into account room or object with a majority vote 
    \item \textbf{Debate:} For each test question, we have the agents of the Multi-LLM system participate in a turn-based discussion where they relay their thoughts to one another for 2 rounds. To initialize the debate we provide each LLM-based agent context with the following system prompt. 
    \\
System prompt: \textit{You are an embodied agent in a house. Your id is [INDEX]. Here are your observations from exploring the house: [OBSERVATION DICTIONARY].  Your initial answer to whether or not there was a [OBJECT] in the [ROOM] was [INITIAL ANSWER]. Other agents may have different answers. Please debate with the other agents to come to a consensus. Here is the conversation history: [CONVERSATION HISTORY]}
    \\
The last sentence is replaced with \textit{This is the beginning of the conversation.} at the start of the debate when no conversation history is present. Below, we provide the user prompt to get the agent's response for its turn during that round. 
\\
User prompt: \textit{It is your turn to speak. Use your observation and conversation history to help.}
\\
When prompted for the final answer at the end of the debate. The user prompt is \textit{Please give your final "Yes/No" answer. Give a definite answer.}
We then collect the final answers from each agent and take the majority vote as the final prediction.
\end{itemize}

\subsubsection{Central Answer Model (CAM)}
To give the final answer to each question, we train a model $f$ to give a final, binary ``Yes/No" output. The model input dimension is $2 + K$, where $K$ is the number of agents. The first two inputs account for the vectorized question in the form of ``[object, room]'', $(o, r)$. The rest of the $K$ inputs come from entering the independent answers from each of the agents chosen for the multi-LLM system, $\cup_{k=1}^K s_k^{(o,r)}$. $f$ then outputs a predicted answer $\hat{y}$ and is then optimized with the ground truth answer $y$. During inference, we input unseen questions and agent answers into the network to output a test-time answer. Due to the modularity of our approach, for each set of independent responses, we train multiple machine learning models to produce multiple central answer models to compare against baseline aggregation approaches.

\textbf{Note:} The classifier is overfitted to a given environment because the Multi-LLM embodied system is only for a specific environment. With our EQA variation, the agent must answer questions on what it has explored in the given environment. This overfitting allows for the personalization of our systems to the household environment. In our potential use case of a user asking a question of whether or not an object is located in a certain area of their environment, the agents should be trained and asked in the same environment.

We produce CAM models with various machine-learning algorithms: 
Neural Network (NN), Random Forest (RF), Decision Tree (DT), XGBoost (XG), RBF SVM (SVM), Linear SVM (SVM-L), and Logistic Regression (LR). The NN was implemented with Pytorch using 3 linear layers. The dimension were 16, 8, 1 in that order with a Sigmoid activation function at the end to ensure the output was in between 0 and 1. XGBoost used the default settings from the xgboost library package. RF from the sklearn library used $1000$ trees, and all other methods used the default sklearn settings.

% We train multiple non-parametric (Random Forest, Decision Tree, XGBoost) and parametric models (Neural Network, Logistic Regression Classifier) as central answer model. 
% We use a sigmoid activation function to output that represents the probability, $p$ that the object is in the room at the given timestep. Letting $y \in {0, 1}$ be the ground truth answer (1 representing ``Yes''), we use the following binary cross-entropy loss to update the network during training,

% \begin{equation}
% L(y, p) = y \log p + (1-y)\log (1-p).
% \end{equation}

% Section \ref{sec:results} highlights the results of our approach over single LLM-based agents, consensus-reaching debate, and majority vote.

\section{Experimentation}\label{sec:exp}

% We now describe our experimentation in detail. With these experiments, we aim to 1) measure the accuracy of CAM for EQA and show its performance over other aggregation baselines (Results section); 2) execute our Multi-LLM framework with observations gathered from a SOTA exploration method for unknown household environments, demonstrating its practicality for real-world scenarios (CAM + Language+Guided Exploration section); and 3) provide a feature importance analysis to highlight the benefit of utilizing independent agent responses in our approach (Feature Importance Analysis Section)
% In all experiments we use GPT-4-turbo.

We now describe our experimentation in detail. With these experiments, we aim to 1) measure the accuracy of CAM for EQA and show its performance over other aggregation baselines (Results section); 2) and 3) provide a feature importance analysis to highlight the benefit of utilizing independent agent responses in our approach (Feature Importance Analysis Section).
In all experiments we use GPT-4-turbo.

% \am{Can  we summarize here what are the specific questions we are going to investigate in the experimental section. In other words, present the intuitive high level reasoning of the experiment plan, like why we did whatever we did for the experiments, and why it makes sense to test those things.}

\subsection{Query Generation}\label{sec:qg}
We extract the embodied question bank from the Matterport3D environments. For each scan, we extract an offline list of observed items detected in each image of the environment using the GLIP object detection algorithm \cite{li2022grounded}. The items space observed by GLIP can be found in the Appendix in Table \ref{tab:label_encodings}.

For each item and room combination, we formulate the question using the template described in Section \ref{sec:qa}. Because these questions were extracted based on the observed items in the scan, we know the ground truth answer is ``Yes".

To create questions with ground-truth answers ``No", for each query generated with a ``Yes" ground-truth, we formulate a question with the same item but with a random room that is not within the subset of rooms that contains that specific item.

\textbf{Note:} Our setup is highly dependent on the object detector used. Error can thus be introduced by the object detection with GLIP and then the LLM answer generation. Some outputs of GLIP included toilets in the living room and a TV in the hallway. When the LLM agent has a (room, item) in its observation memory in the system prompt, it will usually say Yes if asked about said room and item, but when it does not have it in memory, the questions are harder to answer, leading to error in answers generated by the independent LLM agent. More error analysis on agent answers will be studied further in the Section \ref{sec:feat_imp}.

\subsection{Evaluation}

% In this section, to focus on the aggregation abilities of CAM, we have an oracle exploration where we explicitly say which rooms and items an agent has in its observations memory when answering the questions. In Section \ref{sec:camlgx} 
We use a SOTA algorithm, LGX \cite{dorbala2023can} for the exploration phase.
% Given a house environment, over five trials, we randomly split the set of questions into a training and test set. 

\noindent \textbf{Metric.} To measure and compare the performances of our various MELE aggregation methods, we calculate the inference time accuracy, or correctness, of the model:

\begin{equation}\label{eq:correct}
    \frac{1}{N}\sum_{i=1}^N \mathbf{1}_{y_i = \hat{y}_i},
\end{equation}

where $y_i$ is the ground truth label of the test question, and $\hat{y}_i$ is the final answer prediction from the given aggregation method.

\subsection{Language-Guided Exploration}

% \begin{figure*}[!h]
%     \centering
%     \includegraphics[width = 0.9\textwidth]{latex_cam_figures/malicious_agents.png}
%     \caption{\textbf{Malicious;} Performance of CAM with $3$ independent agent where one agent is malicious, contributing responses it believes to be incorrect. We use a 90\%-10\% train-test random split. We experiment on Matterport environments with \textbf{[LEFT]} 215 nodes and \textbf{[RIGHT]} 53 nodes. Against both baselines, all CAM methods outperform during test-time inference, with XGBoost having the highest accuracy. In cases where an agent is malicious, our CAM method is better suited that majority vote and debate.}
%     \label{fig:malicious}
% \end{figure*}

\begin{table*}
    \centering
    \resizebox{\textwidth}{!}{\begin{tabular}{|c|c|c|c|c|c|c|c|c|c|}
        \hline
        & \multicolumn{9}{|c|}{\textbf{Accuracy (\%)}} \\
        \hline
        & \multicolumn{7}{|c|}{\textbf{CAM-based Methods (Ours)}} & \multicolumn{2}{|c|}{\textbf{Baselines}} \\ 
        \hline
        \textbf{Environment Size (num of nodes)} & \textbf{NN} & \textbf{RF}  & \textbf{DT} & \textbf{XG}  & \textbf{SVM} & \textbf{SVM-L}  & \textbf{LR}  & \textbf{MV}  & \textbf{Debate}\\ \hline
         20 & 65.56 & 63.33 & \textbf{73.33} & \textbf{73.33} & 64.44 & 43.33 & 53.33 & 58.89 & 58.89 \\ \hline
         43 & 61.00 & 63.00 & \textbf{86.00} & 82.00 & 62.00 & 60.00 & 64.00 & 41.00 & 40.00 \\ \hline
         53 & 50.83 & 68.33 & 78.75 & \textbf{81.25} & 64.17 & 50.83 & 50.83 & 44.58 & 44.17 \\ \hline
         58 & 50.00 & 74.00 & 84.00 & \textbf{87.00 }& 74.00 & 55.00 & 53.00 & 54.00 & 53.00 \\ \hline
         78 & 65.00 & 70.45 & 85.45 & \textbf{86.36} & 70.45 & 57.27 & 55.91 & 42.27 & 41.82 \\ \hline
         114 & 64.06 & 77.19 & 81.25 & \textbf{87.50} & 70.94 & 55.63 & 60.31 & 51.88 & 47.19\\ \hline
         145 & 65.33 & 75.33 & 83.00 & \textbf{87.67} & 69.00 & 58.33 & 59.33 & 46.67 & 45.33\\ \hline
         215 & 66.33 & 72.67 & 79.67 & \textbf{88.33} & 69.67 & 56.00 & 56.33 & 42.00 & 43.00 \\ \hline
         & \multicolumn{9}{|c|}{\textbf{Accuracy (\%) with Malicious Agent}}\\  \hline
        & \multicolumn{7}{|c|}{\textbf{CAM-based Methods (Ours)}} & \multicolumn{2}{|c|}{\textbf{Baselines}} \\ \hline
        \textbf{Environment Size (num of nodes)} & \textbf{NN} & \textbf{RF}  & \textbf{DT} & \textbf{XG}  & \textbf{SVM} & \textbf{SVM-L}  & \textbf{LR}  & \textbf{MV}  & \textbf{Debate}\\ \hline
        20 & 61.11 & 62.22 & \textbf{73.33} & \textbf{73.33} & 64.44 & 44.44 & 53.33 & 61.11 & 58.89 \\ \hline 
         53 & 60.42 & 68.33 & 78.75 & \textbf{81.25} & 64.17 & 49.58 & 50.83 & 46.67 & 44.17 \\ \hline
         145 & 68.00 & 75.33 & 83.00 & \textbf{87.67} & 69.00 & 59.33 & 59.33 & 47.67 & 45.67\\ \hline
         215 & 64.33 & 73.67 & 79.67 & \textbf{88.33} & 69.67 & 56.33 & 56.33 & 43.67 & 43.00\\ \hline
    \end{tabular}}
    \caption{Performance measured in accuracy (Eq \ref{eq:correct}) of MELE with $3$ independent agent that performed LGX for collecting observations. We use a 90\%-10\% train-test random split. We experiment in $8$ different environments over $5$ trials. Against both baselines, all CAM methods outperform during test-time inference, with XGBoost having the highest accuracy. We repeated the experiment in $4$ of the environments but turned one of the agent malicious by providing responses opposite to what it believes are the correct answers. In cases where one agent is malicious, our CAM method is still better suited than MV and debate.}
    \label{tab:summary_results}
\end{table*}

\begin{figure}[h]
    \centering
    \includegraphics[width = 0.9\columnwidth]{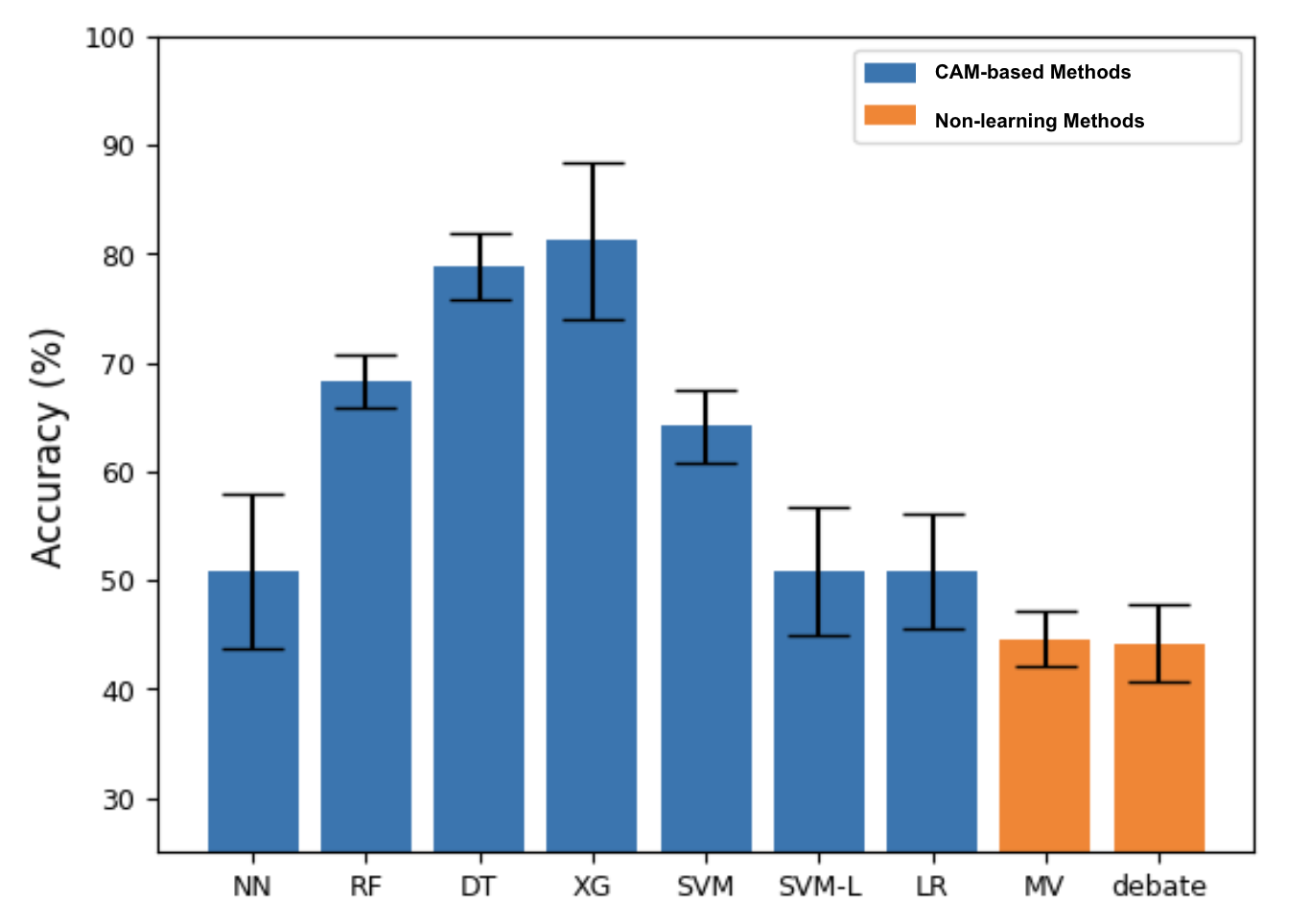}
    \caption{ Performance of MELE methods with $3$ independent agent that performed LGX for collecting observations. We use a 90\%-10\% train-test random split. We experiment on a Matterport environment with 215 nodes. We report the mean and standard deviation over $5$ trials. Against both baselines, all CAM methods outperform during test-time inference, with XGBoost having the highest accuracy.}
    \label{fig:performance_eval_lgx}
\end{figure}

% In this section, we evaluate our Multi-LLM framework by gathering observations using a SOTA exploration method. Rather than manually selecting which observations to provide to the LLM to answer the queries, 
We have all the LLM-based agents explore the household environment for a set amount of time before answering questions. In the exploration phase, each subordinate agent is set in the Matterport house scan graph and performs an augmented variation of Language-Guided Exploration (LGX) as proposed by \citet{dorbala2023can}. In traditional LGX, the LLM agent observes the immediate household items around itself and chooses which one to navigate to find a specific target item, forming a chain of ``hops'' to the item.  

In our variation, we instead provide the LLM with a system prompt that encourages the agent to explore as much of the house as possible. 
\textit{You are an embodied agent in a house. You want to aggressively explore the house and you want to find as many unique objects as you can.}

The agent then travels from node to node in the house graph. For each timestep, we document the timestep, node, and items observed. When the agent reaches a node we can then access the stored results and add them to the list of observed items. We have the $3$ agents perform $10$ steps of this augmented LGX and store the observations. 
The rest of the experiment process follows the same setup as described in Section \ref{sec:pipeline} with the Question-Answering and Aggregation phases. We use a $90\%$-$10\%$ random train-test split over $5$ seeds (0 to 4 are used as the seeds). Experiments were run on an Apple M1 Pro and macOS 14.5.

\textbf{Results:} The performance of the various CAM methods and non-learning aggregation methods in the $8$ graph environments is shown in Table \ref{tab:summary_results}. We test in environments of different sizes (measured in number of nodes) to showcase the generalization of the CAM approach across different size complexities. From the table, we see CAM model consistently outperforms both MV and debating in accuracy (Eq \ref{eq:correct}. We report the average accuracy over $5$ seeds. We see that the non-linear methods consistently achieve higher accuracy than the linear methods, suggesting that the dataset contains no hyperplane separating the queries with ``Yes'' and ``No'' ground-truth answers. Specifically, XGBoost attains the highest accuracy in all but one environment, suggesting that the boosted ensemble method best fits the generated query-answer data. 
% In fact, with LGX for observation collection, we see that on average, all CAM methods have a higher test time accuracy than the non-learning aggregation baselines. 

Interestingly, in all but one of the environments, debating does slightly worse on average than MV. This degradation occurs due to agents who were initially correct being swayed by incorrect agents during conversation. Overall, our Multi-LLM system can be utilized in conjunction with an LLM-based exploration method, emphasizing its practicality in real-world use cases.

\begin{table}
    \centering
    \begin{tabular}{|c|c|c|c|}
    \hline
       Method & \textbf{Agent 1}  & \textbf{Agent 2}  & \textbf{Agent 3} \\ \hline
    
        XG & 38 $\pm$ 8.3 & 38 $\pm$ 7.0 & 38 $\pm$ 6.2\\ \hline
        MV & 97 $\pm$ 0.1 & 95 $\pm$ 0.1 & 96 $\pm$ 1.6 \\ \hline
        debate & 97 $\pm$ 1.4 & 93 $\pm$ 3.1 & 91 $\pm$ 3.6\\ \hline
    \end{tabular}
    \caption{Agreement (\%) between each LGX agent and the final answers of 1) XGBoost-based CAM (XG), 2) Majority Vote (MV) and 3) debate in the $215$-node environment. We report the average and standard deviation over the $5$ trials. We see that XGBoost learns not to agree with any of the agents to achieve its high accuracy, suggesting more exploration or better object detection is needed.}
    \label{tab:agreement}
\end{table}

\textbf{Agreement: } We analyze the agreement between each agent and the XGBoost model (the best performing CAM-based method), majority vote, and debate. We measure agreement between an agent and each central module method by the percentage of similar answers:
\begin{equation}\label{eq:agreement}
    Agr(s_k, f) = \frac{1}{N}\sum_{i=1}^N \mathbf{1}_{s_k^i = \hat{y}_i},
\end{equation}
where $i$ is index of the query parameterized by $(o,r)$. Table \ref{tab:agreement} reports the agreement over $5$ trials and we see that XGBoost has learned not to agree with the agents as much as MV and debate to attain its high accuracy. This low agreement could stem from the errors introduced by the object detection of GLIP and the question-answering of each individual agent. We further analyze the independent answers' impact on training in the Section \ref{sec:feat_imp}.

\begin{table*}[!t]
    \centering
    \begin{tabular}{|c|c|c|c|c|c|}
        \hline
        \textbf{Query Subject} $\downarrow$ & Object & Kitchen & Dining Room & Bedroom & Office  \\
        \hline
        Kitchen & 27.50 & 22.50  & 27.50 & 27.50 & 22.50  \\      
        \hline
        Dining Room   & 17.50 & 12.50 & 12.50 & 17.50  & 12.50   \\ \hline

        Bedroom & 25.00   & 25.00  & 25.00 & 25.00  & 25.00  \\ \hline

        Office & 12.50   & 5.00 & 5.00 & 12.50 & 7.50 \\  \hline

    \end{tabular}
    \caption{This table displays the feature importance of different decision tree-based CAM models. We see in cases where the queries only pertaining to the dining room (third row) that responses based only on dining room observation (fourth column) is not the most important feature. This insight suggests that CAM learned not to rely on that LLM agent.}
    \label{tab:feat_xg}
\end{table*}
\noindent \textbf{Malicious Agent:} We also show the benefits of CAM where one of the three agents is malicious. To turn the agent malicious we invert its independent answers so that all ``Yes'' answers turn to ``No'' and vice versa. We then train all CAM methods and execute MV and debating regularly. Thus, the malicious agent contributes an answer it believes to be incorrect. From Table \ref{tab:summary_results}, we see that our CAM methods still outperform MV and debate. We see that the debate baseline again does worse the majority vote, showing that the incorrect agent was able to sway the overall decision that was initially going to be correct. 

% Interestingly, in the 53-node environment debating baselines do slightly worse on average than MV. This degradation occurs due to agents who were initial lly correct being swayed by incorrect agents during conversation. Overall, our Multi-LLM system can be utilized in conjunction with an LLM-based exploration method, emphasizing its practicality in real-world use cases. In the next section, we provide a feature importance and interpretability analysis of CAM.
% In Section \ref{sec:results}, analyze the effect of these different prompts on the amount of unique node coverage and node visitation frequency.

\section{Feature Importance Analysis}\label{sec:feat_imp}

In the context of machine learning, feature importance is a metric by which input impacts the model's final answer. For our MELE system, feature importance can be used to examine if any agent response or query context (i.e. room or object) is relied heavily upon for training the central answer model. To examine which features hold the most weight in the final decision of the CAM model, we look into studying the variable, or feature importance \cite{breiman2001random, Fisher2018AllMA}, of the CAM model. We measure feature importance using permutation feature importance (PFI). We define the PFI the $i$-th feature as,

\begin{equation}\label{eq:pfi}
    PFI(i) = |a_{\text{val}}(m) - a_{\text{val}}^i(m)|,
\end{equation}

where $a_{\text{val}}(m)$ is the model $m$ accuracy measured by Equation \ref{eq:correct} of the original validation set, and $a_{\text{val}}^i(m)$ is the accuracy with a modified validation set where the $i$-th feature column is randomly permuted. Intuitively, the more important a variable, the more the accuracy should change if the values are randomly shuffled because the variable is highly correlated with the target variable. In our Multi-LLM system for EQA, an agent that gives nearly-deterministic answers (e.g. ``No'' for almost all answers) would have a PFI close to $0$.

To examine feature importance, we train multiple DT-based CAM models, each with a different filtered query dataset from the $215$-node environment. To better highlight and focus on the feature importance aspect, we have an oracle exploration where we explicitly say which rooms and items an agent has in its observations memory when answering the questions. This feature importance analysis could be applied to dynamic agents that move around with LGX.

In each training, the features comprise the room and object name to represent the query and $4$ independent responses from LLMs that had observations either from the kitchen, dining room, bedroom, or office. The dataset however will only consist of rows with a given ``Room'' value.  The model is trained and tested, and we then calculate feature importance using Equation \ref{eq:pfi}. 

Table \ref{tab:feat_xg} summarizes our feature importance findings. The leftmost column specifies the room the dataset was filtered for, and the other columns show the average PFI of the object feature and independent responses for that filtered dataset over 5 trials. We see that in many cases, like when trained only on the dining room queries, the corresponding feature (i.e. response based only on dining room observation) does not have the highest PFI. We analyzed the offline LLM-generated answers based only on dining room observations and out of the $600$ responses, $549$ are ``No'', or $91.5\%$. Thus, the CAM model has learned not to rely on these answers as much as the other responses even though they are based only on observations from the room in question. This outcome demonstrates how analyzing feature importance can help examine individual members of a multi-agent setup for error analysis and further supports our approach utilizing independent responses. Please see the Appendix for a visualization of the Decision Tree to highlight the benefits of the interpretable ML-based CAM methods.

\section{Conclusion}\label{sec:conclusion}

In this paper, we use  Multi-Embodied LLM Explorers (MELE) to understand the effects of embodied exploration of multiple independent LLM-based agents on QA task for unknown environment. While other methods in the field of Mixture of Experts have used multiple LLM-based models for QA tasks, we specifically investigate this paradigm with a dependence on embodied exploration in a household environment. We compare the accuracy of various aggregation methods: majority voting, debating, and CAM. We show that various machine learning methods are suitable to train CAM and consistently outperform the non-learning-based aggregation methods. 

A key benefit of CAM is that it requires no communication during inference, reducing time costs. We also show that CAM is more robust to malicious agents than the majority vote and debate methods. We end this work by providing a feature importance analysis to study the effect of individual agents and target object specification on the final answer. We hope that this work will encourage research in embodied Multi-LLM for QA as LLM-based agents become more household.

\section{Limitations and Future Work}\label{sec:limitations}

This work has several limitations. First, the ground truth labels to train CAM can be difficult to obtain, especially if the household environment is constantly changing with non-stationary items. We leave it as future work to adapt this Multi-LLM setup for EQA in dynamic household environments. Second, our queries are all objective binary ``Yes/No'' questions. To better enhance practicality, future work should look into aggregation methods for situational, subjective, and non-binary questions. It would also be useful to remove the need to explicitly mention the target object in the query. Other areas directions for research could be then to apply our CAM aggregation method to question-answering tasks outside of Embodied AI such as long video understanding. Third, our results hinge on the quality of the observation data. In our work, we relied on the GLIP model. Future work should analyze the effect of scene understanding algorithms on our setup.

\bibliography{main}
\bibliographystyle{latex/acl_natbib}

\newpage
\appendix

\onecolumn

\section{Code and Dataset}

We provide an anonymous link to our code \footnote{\url{https://anonymous.4open.science/r/CAM-embodied-llm-54CE/}}.
The query dataset extracted is also within the submitted supplementary materials folder. 

\section{Multi-LLM vs LLM}
To motivate multiple LLM agents we provide a comparison between 1 agent and 3 agents in Table \ref{tab:num_agents}.

\begin{table}[h]
    \centering
    \begin{tabular}{|c|c|c|}
    \hline
         Environment Size & Accuracy (\%) of 1 Agent  & Accuracy (\%) of 3 Agents (MV)\\ \hline
        $53$ nodes & $41.67$ & $42.00$ \\ \hline
        $215$ node & $44.17$ & $44.58$\\ \hline
    \end{tabular}
    \caption{Rise of accuracy with multiple dynamic agents}
    \label{tab:num_agents}
\end{table}

\section{Additional Visualization of Results}
In this section, to focus on the aggregation abilities of CAM, we have an oracle exploration where we explicitly say which rooms and items an agent has in its observations memory when answering the questions. 

Below we provide additional visualizations of the results of our learning-based CAM methods against MV and debating on other Matterport environments. Like Table \ref{tab:summary_results}, we use independent answers of $3$ LLMs explored with LGX. Majority of the CAM methods consistently outperform the baseline majority vote (MV) and debate methods, with Decision Tree (DT) and XGBoost (XG) being the highest-performing methods.

\begin{figure}[!h]
    \centering
    \includegraphics[width= 0.7\columnwidth]{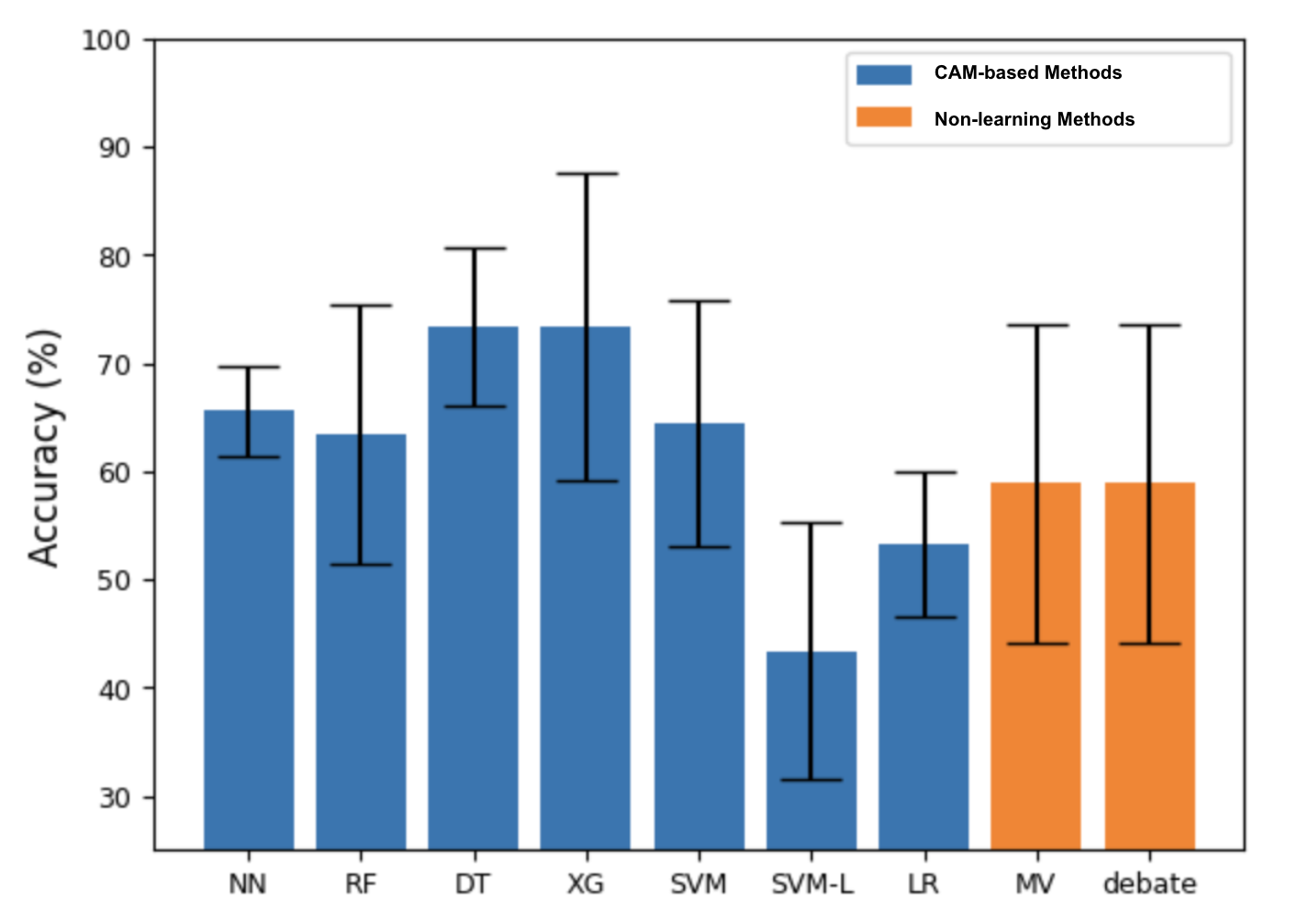}
    \caption{20 nodes}

\end{figure}

\begin{figure}[!h]
    \centering
    \includegraphics[width= 0.7\columnwidth]{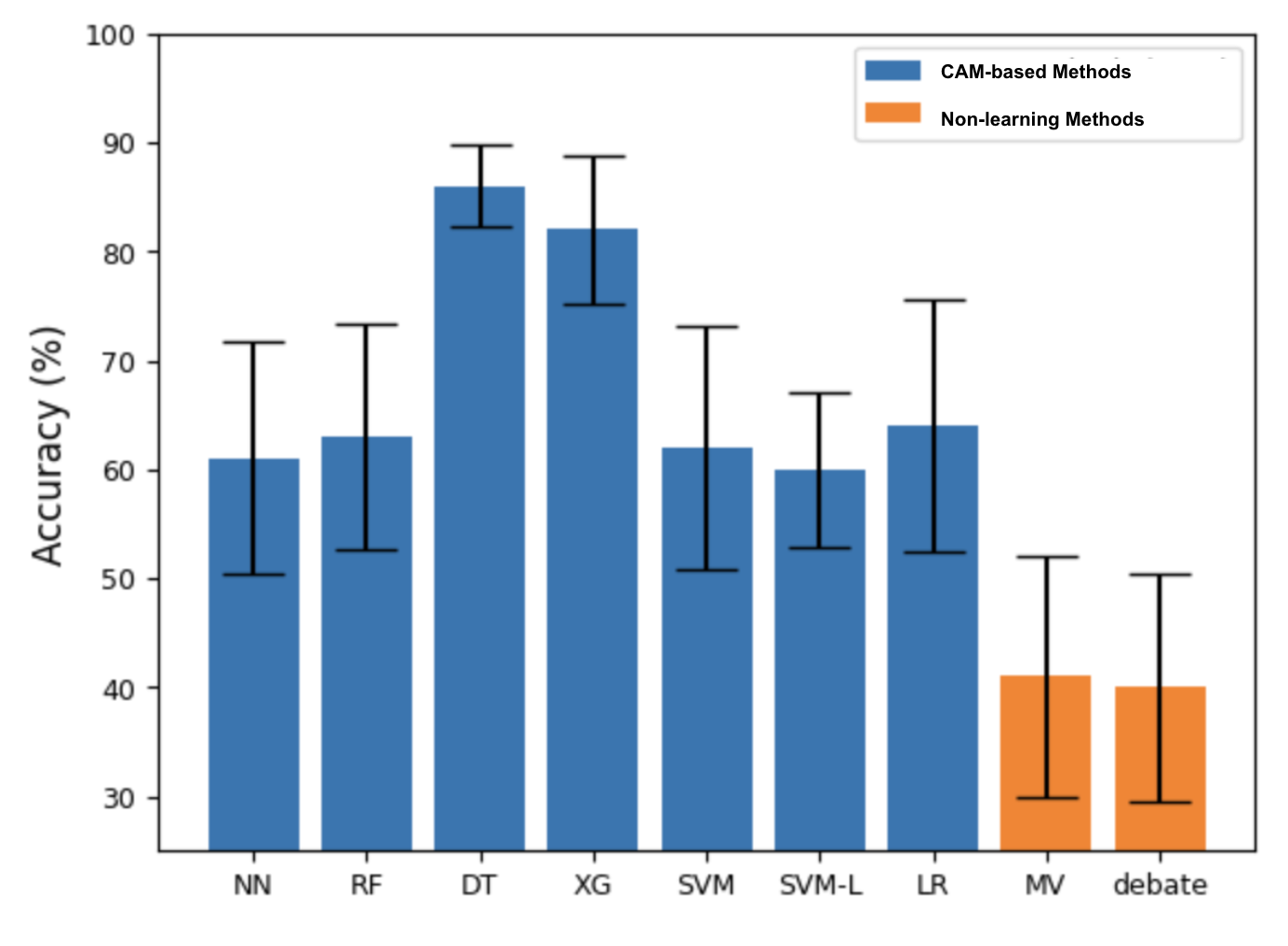}
    \caption{43 nodes}
   
\end{figure}

\begin{figure}[!h]
    \centering
    \includegraphics[width= 0.7\columnwidth]{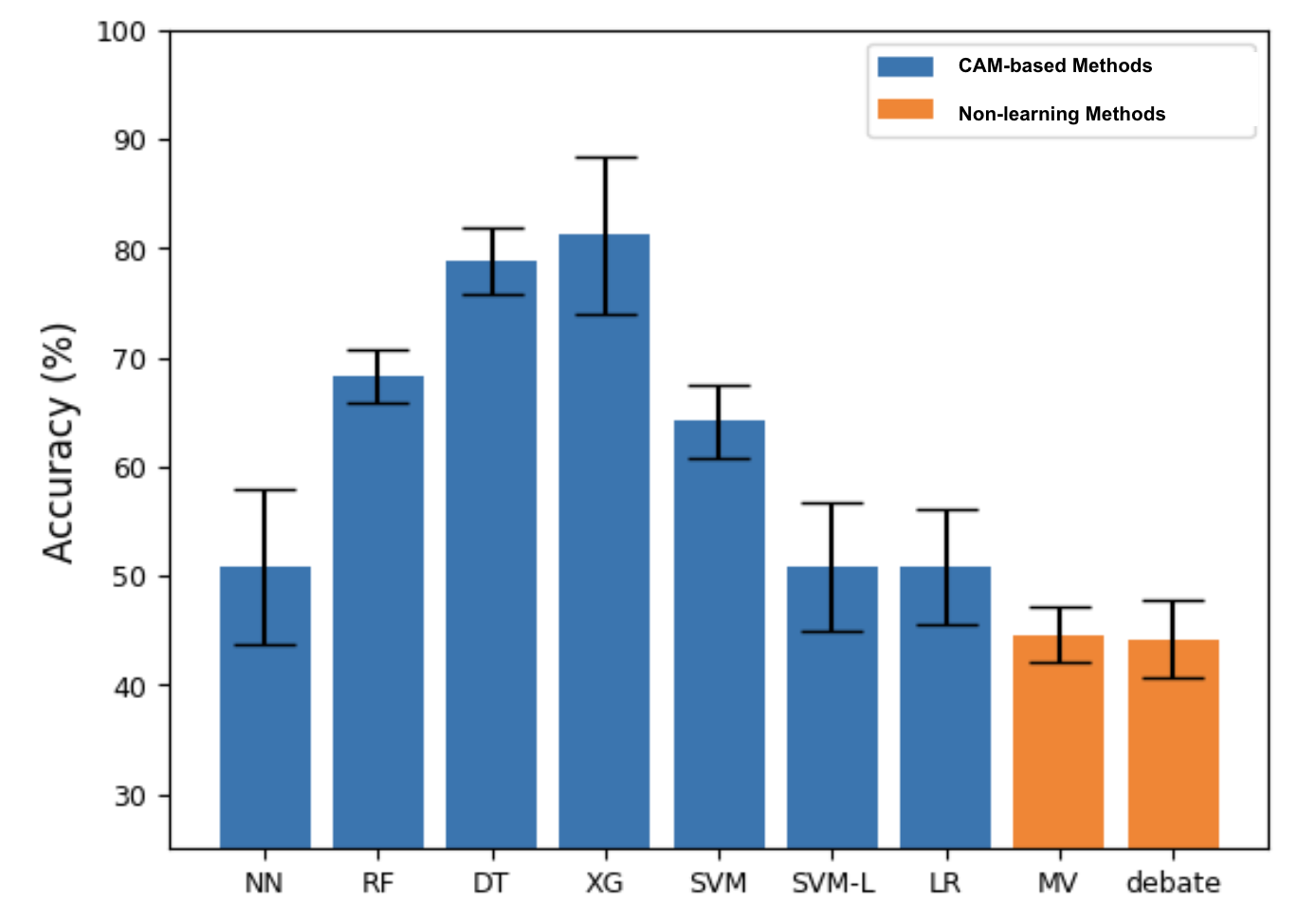}
    \caption{53 nodes}

\end{figure}

% \begin{figure}[!h]
%     \centering
%     \includegraphics[width= 0.7\columnwidth]{latex/figures_appendix/Z6MFQCViBuw.png}
%     \caption{58 nodes}

% \end{figure}

\begin{figure}[!h]
    \centering
    \includegraphics[width= 0.7\columnwidth]{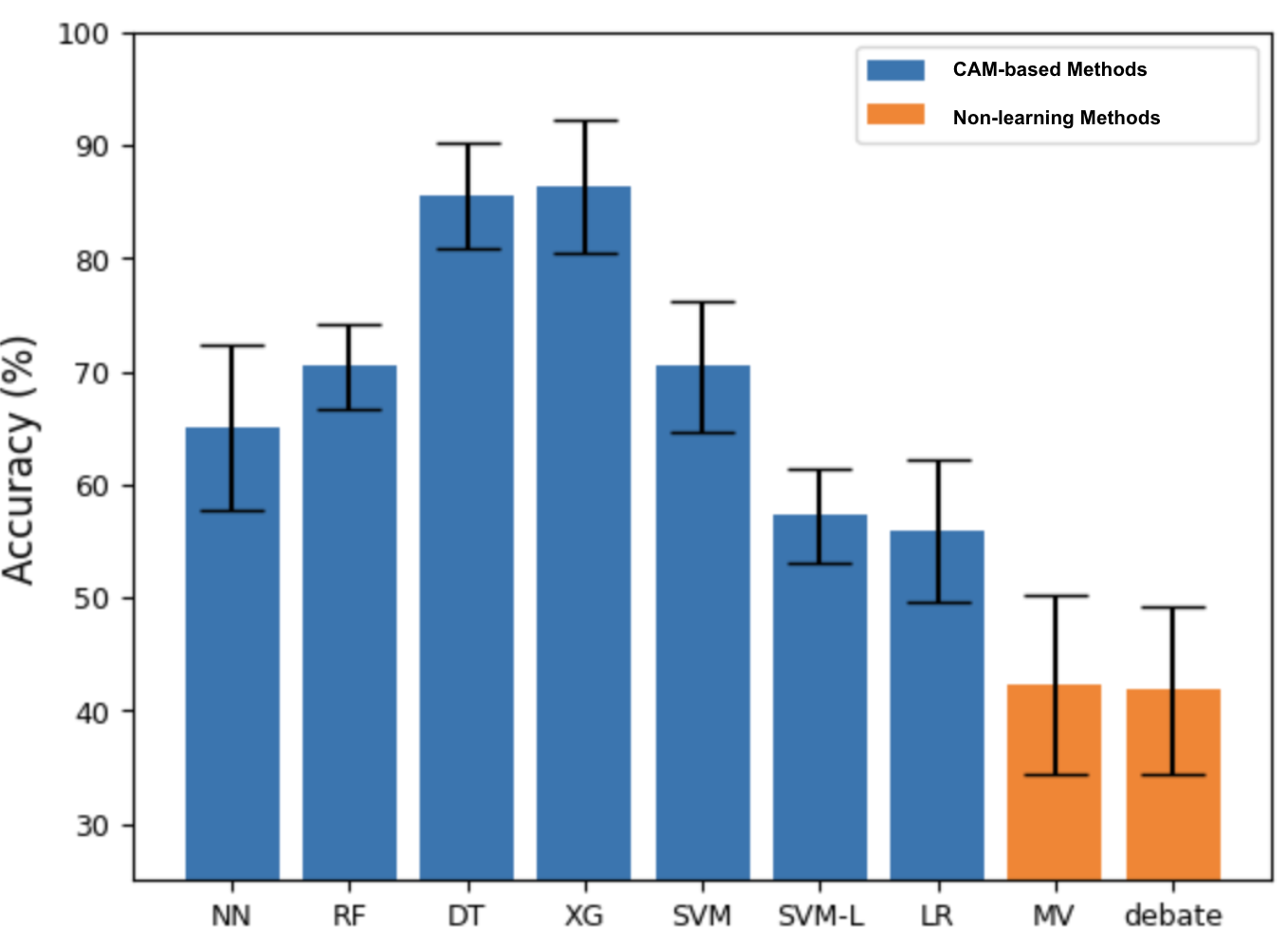}
    \caption{78 nodes}
  
\end{figure}

\begin{figure}[h!]
    \centering
    \includegraphics[width= 0.7\columnwidth]{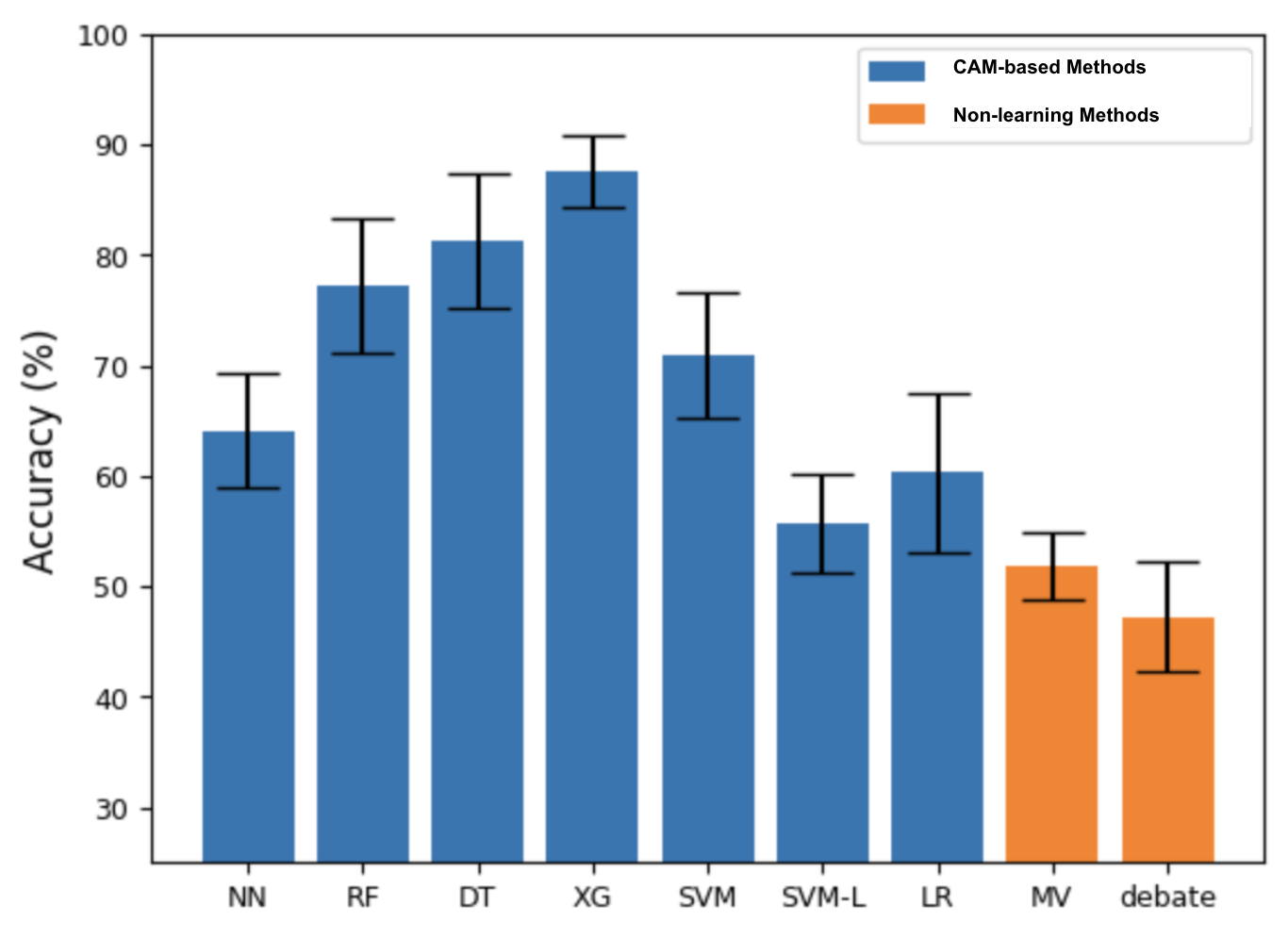}
    \caption{114 nodes}

\end{figure}

\begin{figure}[h!]
    \centering
    \includegraphics[width= 0.7\columnwidth]{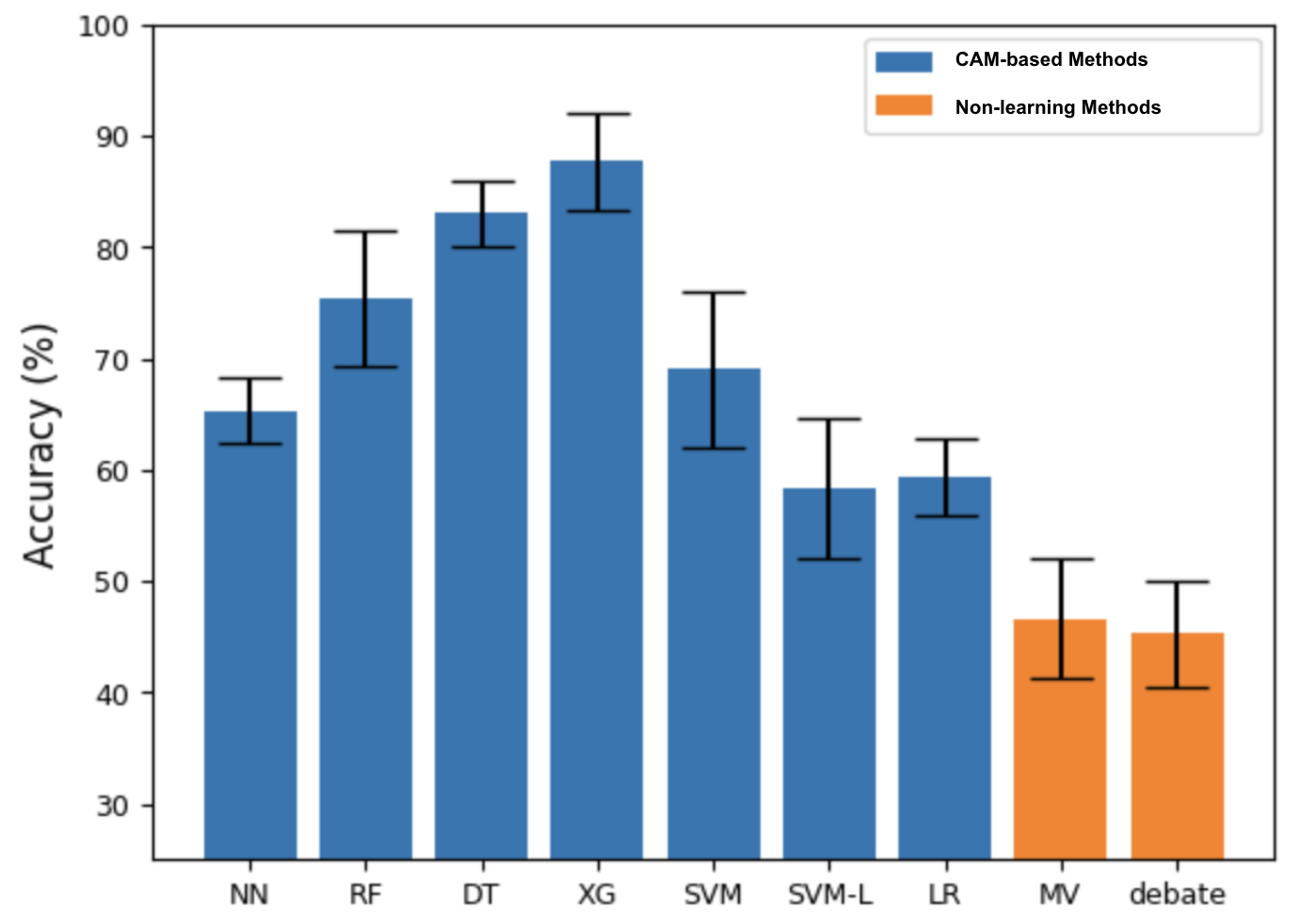}
    \caption{145 nodes}
    
\end{figure}
\newpage

\begin{figure}[!h]
    \centering
    \includegraphics[width = 0.9\columnwidth]{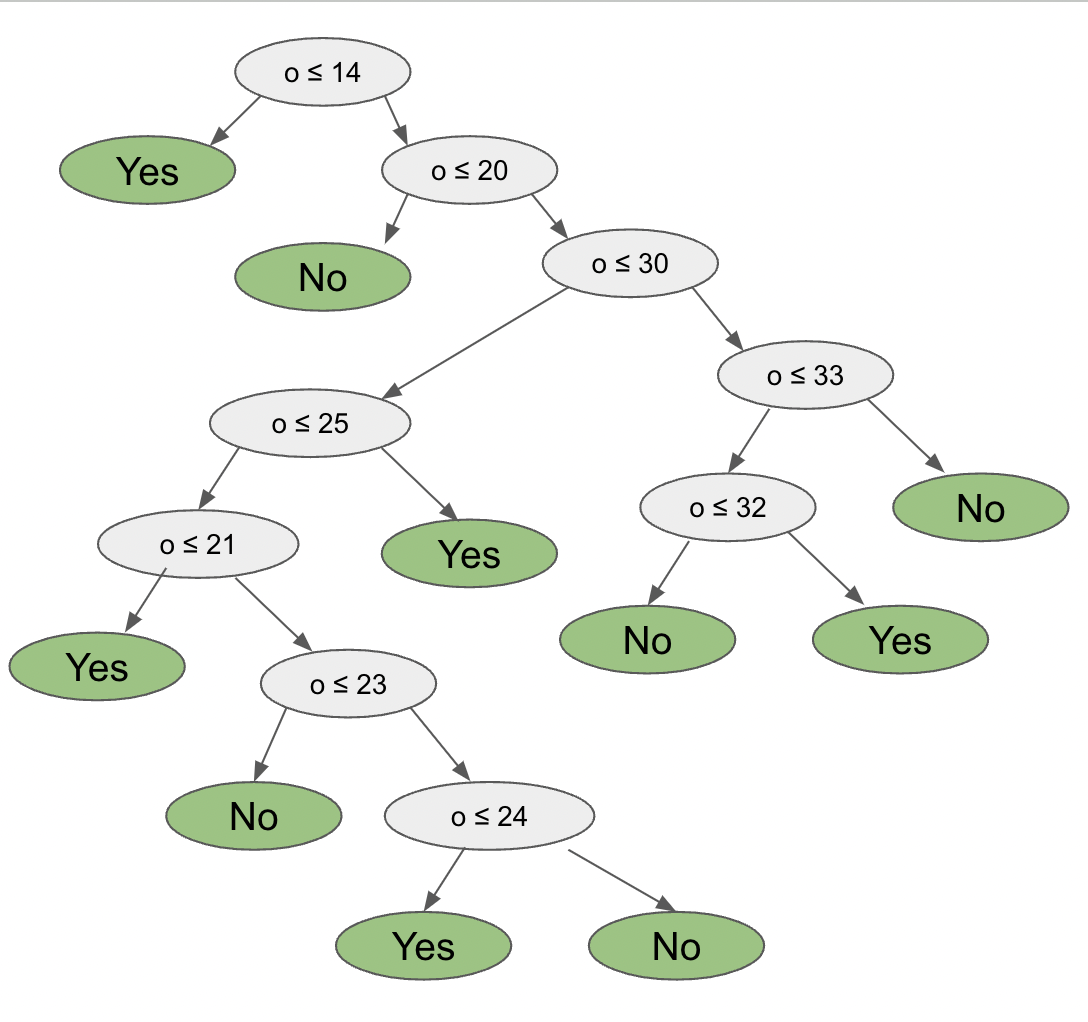}
    \caption{Visualization of Decision Tree from Section \ref{sec:feat_imp}.}
    \label{fig:tree}
\end{figure}

\section{Decision Tree Visualization}

In this section, we showcase the potential interpretability of our learning-based CAM framework by visualizing the decision tree trained in Section \ref{sec:feat_imp}. This tree in Figure \ref{fig:tree} was trained on queries only pertaining to the kitchen and was trained on independent answers from four different LLM agents. One agent had only observations from kitchen; another only dining room; another bedroom; and the last one only had office observations. We see that the main feature used was actually the object query. We provide the object encoding labels in Table \ref{tab:label_encodings}.

\begin{table}[!h]
    \centering
    \begin{tabular}{|c|c|}
        \hline
        \textbf{Object Label} & \textbf{Encoding $o$} \\ \hline
    appliance & 0 \\
    armchair & 1 \\
    bathtub & 2 \\
    bed & 3 \\
    board & 4 \\
    bookcase & 5 \\
    books & 6 \\
    cabinet & 7 \\
    chair & 8 \\
    clothes & 9 \\
    counter & 10 \\
    curtain & 11 \\
    cushion & 12 \\
    desk & 13 \\
    dresser & 14 \\
    dumbells & 15 \\
    hairbrush & 16 \\
    headphones & 17 \\
    jumprope & 18 \\
    mirror & 19 \\
    mug & 20 \\
    nightstand & 21 \\
    phone & 22 \\
    picture & 23 \\
    plant & 24 \\
    playing cards & 25 \\
    refrigerator & 26 \\
    sink & 27 \\
    sofa & 28 \\
    table & 29 \\
    television & 30 \\
    toilet & 31 \\
    toothbrush & 32 \\
    towel & 33 \\
    tv remote & 34 \\
    wallet & 35 \\
    water bottle & 36 \\
    whiteboard & 37 \\
    window & 38 \\
    wristwatch & 39 \\
    \hline
    \end{tabular}
    \caption{Mapping of object labels to encodings $o$}
    \label{tab:label_encodings}
\end{table}

\end{document}